# کنترل فازی پاداش یادگیری تقویتی برای تشخیص ارقام دست‌نویس

صابر ملک زاده[1]

1- گروه علوم کامپیوتر، دانشگاه ولیعصر(عج) رفسنجان، رفسنجان، ایران

**خلاصه**

تشخیص محیط‌های انسانی توسط سامانه‌های رایانه‌ای، همواره یکی از مهم‌ترین موضوعات در دانش هوش مصنوعی بوده است. در این میان تشخیص دستخط انسان و مفهوم‌سازی آن برای رایانه به عنوان یکی از مباحث در این عرصه مطرح شده است. در سال‌های اخیر با گسترش روش‌های یادگیری ماشین در هوش مصنوعی، تلاش برای استفاده از این روش برای حل مسائل مختلف مطرح در هوش مصنوعی افزایش چشمگیری داشته است.

در این مقاله سعی شده با استفاده از کنترل‌کننده‌ای فازی، میزان پاداش یادگیری تقویتی، جهت تشخیص ارقام دست‌نویس بهینه شود. برای این منظور ابتدا یک نمونه از هر رقم با 10 فونت رایانه‌ای معیار، به عامل داده‌شده و سپس به‌صورت مرحله به مرحله به آموزش عامل پرداخته شده است. در مرحله‌ی بعد به آزمایش نمونه‌ها به‌وسیله داده‌های آموزشی پرداخته شده و نتایج به‌دست‌آمده حاکی از تأثیر مستقیم بهبود عملکرد تشخیص، در زمان کنترل فازی میزان پاداش یادگیری تقویتی بوده است.

**واژه های کلیدی:** یادگیری تقویتی، کنترل‌کننده‌ی فازی، تشخیص رقم، دست‌نویس

## 1. مقدمه

یادگیری تقویتی[*] [1] در زمره‌ی روش‌های یادگیری ماشین است که با روش پاداش‌دهی[†] و تقویت عامل با آموزش خوب و بد به آن، عامل را به سمت هدف خوب رهنمون می‌شود. پاداش در هر مرحله به ازای انتخاب درست توسط عامل، می‌تواند به‌صورت دودویی[‡] یا به‌صورت فازی[§] باشد. در سامانه‌ی دودویی به ازای هر عمل مثبت، عامل یک امتیاز مثبت دریافت می‌کند، اما در سامانه‌ی فازی، کیفیت عمل خوب نیز مهم است و عددی بین 0 و 1 به‌عنوان پاداش اختصاص می‌یابد.

در اوایل قرن 20 میلادی ثورندایک[**] [1] ، برآن شد تا با استفاده از آزمایشی، اثر یادگیری تقویتی بر روی مغز انسان را اثبات کند. در این آزمایش گربه‌ای در قفسی کارتونی قرار داده شد. هر بار که این گربه برای بیرون رفتن از قفس تلاش‌های بهتری انجام می‌داد، غذا دریافت می‌کرد. بدین ترتیب گربه (عامل)[††] تلاش‌های خود را به سمت بهترشدن متمرکز کرد تا غذا دریافت کند و پس از چند تلاش که در اواخر، همگی تلاش‌های خوب بودند، توانست از قفس نجات

---

[*] Reinforcement Learning
[†] Reward giving
[‡] Binary
[§] Fuzzy
[**] Edward Thorndike
[††] Actor

پیدا کند. این همان مکانیسمی[*] بود که در مغز انسان نیز وجود داشت. در اوایل دهه‌ی 60 میلادی دانشمندان علوم کامپیوتر سعی نمودند تا این مکانیسم را وارد هوش مصنوعی سازند و با پاداش و جزا، عامل هوش مصنوعی را تقویت کنند تا نتایج را بهینه کرده و سریع‌تر به دست آورند.

در سال 1965 لطفی‌زاده [1,2] مجموعه‌های فازی را معرفی نمود. با ورود منطق فازی به علوم کامپیوتر دانشمندان سعی کردند تا از منطق فازی و در نتیجه‌ی آن کنترل‌کننده‌های فازی در کنترل نتایج و همچنین به‌دست آوردن نتایج از سامانه‌ی دودویی جهت دقت بیشتر استفاده کنند.

در این مقاله به بررسی کنترل فازی پاداش‌های یادگیری تقویتی، زمانی که عامل به سمت انتخاب‌های درست‌تر پیش می‌رود، پرداخته شده است. در الگوریتم یادگیری تقویتی نیز برای تعیین تأثیر نمونه‌های آموزشی، از منطق فازی استفاده‌شده است.

## 2. کارهای مشابه

درگذشته نیز فعالیت‌هایی در این زمینه بوسیله‌ی شبکه‌های عصبی مصنوعی انجام شده است. در این پژوهش‌ها شبکه‌های عصبی که به طور مناسب تنظیم شده بودند توانستند، با درصد بسیار بالایی تشخیص دهند. همچنین از یادگیری تقویتی برای بهینه نمودن تشخیص استفاده شده است که صرفا محدود به پاداش‌هایی با مقادیر ثابت بوده و یا کلاسیک[†] (غیرفازی) بوده است.

## 3. نمونه‌های آموزشی

جهت آموزش دادن[‡] عامل برای یادگیری بهتر و مناسب‌تر نیاز به یک نمونه‌ی آموزشی مناسب بود که تصاویر آن، هم ازلحاظ وضوح مناسب باشند، هم ازلحاظ کیفیت نمونه‌ها و شباهتشان به نمونه‌های واقعی در سطح قابل قبولی باشند و هم تعداد پیکسل‌های کمتری داشته باشند تا عملیات پردازش بر روی ساختار آن‌ها به‌سرعت انجام شود. برای این منظور نمونه‌هایی از ارقام پایگاه داده‌ی MNIST [3] استخراج شدند. این نمونه‌ها، تصاویری با پس‌زمینه‌ی سیاه و نوشته‌های سفید به‌صورت دست‌نویس از ارقام بودند. این پایگاه داده به‌عنوان استانداردی از نمونه‌های آموزشی و آزمایشی از دست‌خط انگلیسی شناخته می‌شود.

## 4. یادگیری

جهت استفاده از یادگیری تقویتی نیاز بود که یادگیری عامل در مراحل مختلف و گام‌به‌گام صورت گیرد. به همـین منظور برای هر یک از ارقام، چند نمونه فونت‌های[§] معیار رایانه‌ای در سایز 28 در 28 پیکسل[**] (به‌اندازه نمونه‌های آموزشی) به همان شکل نمونه‌های آموزشی (با پس‌زمینه سیاه و نوشته سفید) ساخته شد. سپس 80 درصد (بـه صـورت تصـادفی) نمونه‌های آموزشی که شامل 800 نمونه برای هر رقم بودند، با نمونه فونت‌های معیار رایانه‌ای، بـه‌وسـیله‌ی تـابعSSim[††] [1,4] در نرم‌افزار متلب مقایسه شدند.

---

[*] Mechanisms
[†] Classic
[‡] Train
[§] Font
[**] Pixel
[††] Structural Similarity

$$\mathrm{SSIM}(x,y) = \frac{(2\mu_x\mu_y + c_1)(2\sigma_{xy} + c_2)}{(\mu_x^2 + \mu_y^2 + c_1)(\sigma_x^2 + \sigma_y^2 + c_2)}$$

در این فرمول x و y تصاویر ورودی به تابع جهت بررسی شباهت، μ میانگین مقادیر پیکسل‌های تصاویر و σ انحراف معیار مقادیر پیکسل‌های تصاویر است. همچنین ضریب $\sigma_{xy}$ نیز انحراف معیار پیکسل های تصاویر x و y نسبت به یکدیگر است. این معیار مشخص می‌کند که پیکسل در حال بررسی در تصویر x چقدر متفاوت‌تر از همان پیکسل در تصویر y است. مقدار c نیز، ضرایبی جهت ایجاد ثبات در بخش‌های مختلف فرمول هستند که به ایجاد ثبات در فرمول، به هنگام ظهور مخرج ضعیف کمک می‌کنند. این مقادیر به محدوده‌ی مقادیر پیکسل‌ها بستگی دارند.

تابع SSim تشابه ساختاری میان دو تصویر را به‌وسیله‌ی فرمول بالا سنجیده و نتیجه‌ی تشابه را در قالب عددی بین ۱- و ۱ ارائه می‌دهد که به هر میزان این عدد بیشتر باشد، تصاویر داده‌شده به هم شبیه‌ترند. این تابع همان‌طور که در فرمول بالا نیز مشاهده می‌شود، از مقایسه‌ی میان میانگین مقادیر پیکسل‌های دو تصویر با یکدیگر، جهت تشخیص میزان شباهت، استفاده می‌کند. همچنین در نمونه‌های تازه منتشرشده برای این تابع، تفاوت میان دو تصویر با قطعه‌بندی دو تصویر و شباهت‌سنجی قطعه‌های آن‌ها با یکدیگر نیز سنجیده می‌شود. این تابع در سال ۲۰۱۵ برای اولین بار وارد ساختار نرم‌افزار متلب[*] شد.

درنتیجه‌ی مقایسه‌ها، مشاهده شد که تابع SSim برای ارقام مختلف ماکزیمم‌های متفاوتی ارائه می‌دهد. برای مثال تعداد نمونه‌های رقم ۲ که درصد تشابهشان با نمونه فونت‌های معیار رایانه‌ای بیشتر از ۴۰ درصد بود، ۱۰ نمونه بودند، درحالی‌که همین مثال در رقم ۷ تنها ۱ نمونه نتیجه داشت که نشان از نامتوازن بودن پایگاه داده‌ی آموزشی نسبت به تابع SSim بود. بنابراین در آموزش اولیه، ابتدا نمونه‌هایی که درصد تشابه آن‌ها با نمونه فونت‌های رایانه‌ای بیشتر از ۴۰ درصد بودند، انتخاب شدند. سپس همین عملیات با نمونه‌های آموزشی سابق (بدون نمونه‌های انتخاب‌شده به‌عنوان نمونه‌ی خوب در مرحله قبل) انجام شد. در این مرحله نمونه‌های آموزشی با نمونه‌های فونت رایانه‌ای و دست‌نویس‌هایی که به‌عنوان نمونه انتخاب‌شده بودند، مقایسه شدند و از تشابه نمونه‌ی آموزشی با نمونه‌های خوب میانگین گرفته شد و باز نمونه‌هایی که درصد تشابه آن‌ها با تمامی نمونه‌های خوب (نمونه‌ی فونت‌های رایانه‌ای و نمونه‌های خوب مراحل قبل) بیشتر از ۴۰ درصد بود انتخاب شدند. این تکرار تا ۵ مرحله ادامه یافت. زیرا کمتر از آن باعث انتخاب کم نمونه‌های خوب و عدم وجود اطلاعات کافی برای تشخیص مناسب و بیشتر از آن علاوه بر فزونی نمونه‌های خوب و کم شدن سرعت نرم‌افزار در هنگام آزمایش، باعث افزایش نمونه‌های بد در میان نمونه‌های خوب ارقام دارای نمونه‌های خوب بیشتر و درنتیجه باعث تشخیص نامناسب می‌شد.

در تمامی مراحل میزان شباهت نمونه‌ی انتخاب‌شده (تنها) با نمونه فونت‌های رایانه‌ای (حتی آن نمونه‌هایی که در مرحله پنجم انتخاب شدند برای دریافت میزان شباهت، جهت ذخیره، تنها با نمونه فونت‌های رایانه‌ای مقایسه می‌شدند.) به ترتیب شماره‌ی ذخیره‌ی آن در میان نمونه‌های خوب، در فایل‌هایی که برای هر رقم مجزا بودند، ذخیره شدند تا در مرحله‌ی آزمایش میزان شباهت آن‌ها به فونت‌های رایانه‌ای در میزان تأثیرگذاری آن‌ها در پاداش یادگیری تقویتی، مؤثر باشد.

## ۵. آزمایش

جهت انجام مرحله‌ی آزمایش نیاز بود، برای ۲۰ درصد باقی مانده‌ی نمونه‌ها، عملیات شباهت سنجی SSim با نمونه‌های خوب انجام پذیرد. در این میان، میزان شباهت هر نمونه‌ی خوب، به نمونه‌های فونت استاندارد رایانه‌ای

---

[*] Matlab

موردبررسی قرار گرفت تا میزان استاندارد بودن آن نسبت به فونت‌های معیار رایانه‌ای درنتیجه‌ی نهایی مؤثر باشد و هرچقدر نمونه‌ی دست‌نوشته‌ی خوب به فونت‌های رایانه‌ای نزدیک‌تر باشد، اثر آن بر نتیجه‌ی تشخیص نیز بیشتر باشد. میانگین شباهت نمونه‌های خوب به فونت‌های معیار رایانه‌ای متفاوت بود و این بر نتیجه‌ی تشخیص تأثیر می‌گذاشت، به‌گونه‌ای که هر میزان این میانگین برای رقمی بیشتر بود، تمامی نمونه‌های آزمایشی، تمایل بیشتری را به انتخاب آن رقم به‌عنوان رقم نتیجه داشتند. کمترین میزان شباهت نمونه‌های خوب به فونت‌های معیار در هر رقم، از بیشترین میزان آن کم شد و میزان نزدیکی هر عدد به بزرگ‌ترین یا کوچک‌ترین مقدار به‌عنوان مقدار عددی بین ۰ و ۱ انتخاب شد. برای مثال اگر بیشترین میزان میانگین شباهت فونت خوب به فونت‌های معیار ۰.۳۶ و میزان کمترین شباهت ۰.۳۲ بود، عدد اختصاص‌یافته به نمونه‌ی دارای میزان تشابه ۰.۳۲ که کوچک‌ترین مقدار بازه ۰.۲۵، برای ۰.۳۶ که بیشترین مقدار بود، ۰.۷۵ و برای ۰.۳۴ که در میانه‌ی بازه بود ۰.۵ اختصاص یافت. درصورتی‌که کمترین مقدار را برابر صفر و بیشترین مقدار برابر ۱ قرار می‌گرفت، نمونه‌ی خوب دارای میزان تشابه ۰.۳۲ بی‌تأثیر می‌شد و همچنین نمونه‌ی خوب دارای میزان تشابه ۰.۳۶ کاملاً مؤثر بود. درحالی‌که میزان شباهت ۱ تنها مختص نمونه‌ی فونت‌های رایانه‌ای (به طور کامل شبیه به خودش) بود.

این بخش به‌نوعی پیاده‌سازی کنترل‌کننده‌ی فازی بود. چرا که هر زمان میزان تشابه نمونه‌ی خوب به نمونه‌ی معیار رایانه‌ای بیشتر بود، میزان اثرگذاری آن نیز کنترل‌شده و به طبع میزان تشابه نمونه‌ی آزمایشی به رقم مورد بررسی نیز بیشتر بود. این عمل باعث می‌شد تا میزان اثرگذاری نمونه‌های خوب که به‌عنوان پاداش یادگیری تقویتی در نظر گرفته‌شده بودند، به‌جای ۰ یا ۱ بودن، عددی میان ۰ و ۱ بنابه نتیجه‌ی کنترل‌کننده‌ی فازی باشند.

همچنین با توجه به فرمول یادگیری تقویتی:

$$R = \sum_{t=0}^{\infty} \gamma^t r_{t+1},$$

که در این فرمول R میزان پاداش کل، γ عامل تخفیف و r میزان پاداش در مرحله‌ی t+1 از یادگیری است، عامل تخفیف نیز بر میزان اثرگذاری پاداش‌ها مؤثر بود. این میزان اثرگذاری همان میزان تشابه نمونه‌ی خوب به نمونه‌های معیار بود.

درصورتی‌که میزان تشابه نمونه‌های خوب در هر مرحله از یادگیری نسبت به‌تمامی نمونه‌های خوب انتخاب شده در مراحل قبلی به‌عنوان پاداش نمونه‌ی خوب موردبررسی ثبت می‌شد، محتمل بود نمونه‌ای که در مرحله‌ی آخر یادگیری انتخاب‌شده بود میزان پاداش بیشتری نسبت به نمونه‌ی بهتری که در مرحله نخست یادگیری انتخاب‌شده بود، داشته باشد. اما درصورتی‌که پاداش میزان شباهت نمونه‌ی خوب در حال انتخاب، تنها با نمونه‌ی معیار تعیین می‌شد، نمونه‌هایی که در مراحل بالاتر یادگیری انتخاب می‌شدند، نسبت به نمونه‌هایی که در مرحله نخست یادگیری انتخاب‌شده بودند، دارای پاداش کمتری بودند و (در این مثال خاص) این‌همان نرخ یادگیری و همچنین عامل تخفیف مورد استفاده در یادگیری تقویتی است. نرخ یادگیری تعیین می‌کند که نمونه‌های قدیمی به چه میزان به نمونه‌های جدید ترجیح داده شوند و عامل تخفیف نیز تعیین کننده‌ی میزان اثرگذاری هر نمونه است. با توجه به لزوم بیشتر بودن تاثیر نمونه‌های انتخاب شده در مراحل جدید نسبت به نمونه‌های انتخاب شده در مراحل قبلی، نرخ یادگیری و عامل تخفیف هر دو یکسان هستند.

یکی دیگر از مزیت‌های کنترل فازی پاداش یادگیری تقویتی، تعیین میزان خوب یا بد بودن تشخیص است. برای مثال اگر تفاوت میان رقمی که دارای بیشینه مقدار به‌دست‌آمده در تشخیص یک نمونه‌ی آزمایشی است و نمونه‌ی بیشینه‌ی بعد از آن بسیار کم باشد، می‌توان حدس زد به احتمال زیاد تشخیص صحیح نبوده است. زیرا مقدار شباهت نمونه‌ی آزمایشی نزدیک به هر دو رقم دارای بیشینه مقدار است. در این صورت می‌دانیم که انتخاب صحیح به طور حتم یکی از این دو رقم خواهد بود ولی انتخاب اینکه کدام‌یک از آن ارقام، انتخاب صحیح هستند، کمی مشکل است و این مشکل در تشخیص را می‌توان به کاربر استفاده‌کننده، اعلام نمود.

## 6. نتایج

در آزمایشی که بر روی ۲۰ درصد نمونه‌های پایگاه داده (۲۰۰ نمونه) از هرکدام از ارقام انجام شد، نتایج زیر به دست آمد:

- بدون استفاده از یادگیری تقویتی و کنترل‌کننده‌ی فازی (تنها با استفاده از تابع SSim) نتیجه‌ی به دست آمده، تشابه ۴۸ درصدی بود و باز از نظر زمانی نیز قابل عملیاتی شدن در نمونه‌های تجاری نبود.
- بدون استفاده از یادگیری تقویتی و با استفاده از کنترل‌کننده‌ی فازی (مقایسه با تمام نمونه‌های یادگیری بدون انتخاب بهترین آن‌ها با استفاده از یادگیری تقویتی و ضرب تشابه بدست‌آمده در تشابه نمونه‌ی یادگیری نسبت به نمونه‌ی معیار) نتیجه‌ی تشخیص ۵۳ درصد بود. با توجه به زمان زیاد لازم برای مقایسه، از نظر زمانی نیز قابل عملیاتی شدن در نمونه‌های تجاری نبود.
- بدون استفاده از کنترل‌کننده‌ی فازی (بدون استفاده از ضریب تشابه میان نمونه‌های خوب و نمونه‌های معیار رایانه‌ای) و با استفاده از یادگیری تقویتی، این درصد تشخیص به ۷۲ درصد کاهش پیدا نمود.
- به‌وسیله‌ی شیوه‌ی ارائه شده در این مقاله، نتایج به دست آمده به طور میانگین، تشخیص ۸۸ درصدی را نشان می‌دادند.

از مهم‌ترین روش‌های تشخیص ارقام دست‌نویس، استفاده از شبکه‌های عصبی مصنوعی عمیق است. در این روش توابعی برای تشخیص ساختار ارقام دست نویس در لایه‌ی پنهان شبکه‌ی عصبی قرار داده می‌شود تا ساختار نمونه را در سطوح مختلف (از جزئیات تا کلیات) بررسی کرده و میزان تشابه ساختاری دو نمونه را بازگرداند. این روش به تازگی نتایج خوبی به دست داده و توانسته تشخیص بالای ۹۵ درصد را نیز ارائه کند.

هرچند شبکه‌های عصبی مصنوعی درصد تشخیص بالایی را ارائه می‌دهند [5] ، اما در این مقاله سعی شده است تا از تابع SSim جهت تشخیص میزان تشابه استفاده شود. با توجه به اینکه استفاده از شبکه های عصبی مصنوعی موجب ایجاد تفاوت در موضوع مقاله نمی‌شد، به دلیل سادگی بیان مسئله از تابع SSim که یکی از بهترین توابع تشخیص آماری شباهت دو تصویر نیز بود، استفاده شد. با استفاده از این تابع، کنترل فازی پاداش یادگیری تقویتی، باعث ایجاد بهبود محسوس‌تری در نتایج حاصله، نسبت به استفاده از شبکه‌های عصبی مصنوعی می‌شد که می‌توانست بهبود درصد تشخیص را به خوبی نشان دهد. البته این بدان معنا نیست که در صورت استفاده از شبکه‌های عصبی مصنوعی به جای SSim و کنترل فازی پاداش یادگیری تقویتی، بهبودی در نتایج حاصل نشد، بلکه تنها میزان بهبود درصد تشخیص ارقام، کمتر بود.

## 7. نتیجه گیری

نتایج به‌دست‌آمده حاکی از آن بودند که استفاده از یادگیری تقویتی تأثیر بیشتری نسبت به کنترل‌کننده‌ی فازی درنتیجه به‌دست‌آمده دارد، اما این بدین معنی نیست که کنترل‌کننده‌ی فازی باعث بهبود نتایج به‌دست‌آمده نمی‌شود. استفاده از ترکیب هر دو روش به‌وضوح نشان‌دهنده‌ی بهبود قابل‌توجهی در نتایج الگوریتم SSim بود و این همان هدف مقاله است.

شاید مهم‌ترین کاربرد این روش را بتوان در تشخیص دست‌خط در زبان‌های مختلف و تبدیل آن‌ها به نوشته‌های رایانه‌ای تعیین کرد. البته با توسعه‌ی این ترکیب بر روی نمونه‌هایی از جنس‌های مختلف، می‌توان در بسیاری از موارد به بهبود نتایج تشخیص یا انتخاب، کمک کرد.

ازجمله کارهایی که در این زمینه در آینده قابل اجرا است، می‌توان به استفاده از ترکیب یادگیری تقویتی و منطق فازی در بهبود نتایج الگوریتم شبکه‌های عصبی بر روی تشخیص دست‌خط پرداخت. همچنین در سال‌های اخیر استفاده از روش‌های یادگیری عمیق بر بستر شبکه‌های عصبی مصنوعی نیز گسترش‌یافته‌اند که با در نظر گرفتن این موضوع تلاش

برای بهبود الگوریتم‌های تشخیص دست‌خط به‌وسیله تشخیص ساختاری تصاویر در لایه‌های پنهان و عمیق شبکه‌های عصبی مصنوعی و ترکیب با یادگیری تقویتی و تعیین پاداش‌های یادگیری به‌وسیله‌ی کنترل‌کننده‌های فازی می‌تواند به بهبود هر چه بیشتر نتایج کمک شایانی کند.

مراجع